# Transfer learning-based method for automated e-waste recycling in smart cities

Nermeen Abou Baker[1,*], Paul Szabo-Müller[1] and Uwe Handmann[1]

[1]Ruhr West University of Applied Sciences, Lützowstr. 5, 46236 Bottrop, Germany

## Abstract

INTRODUCTION: Sorting a huge stream of waste accurately within a short period can be done with the support of digitalization, particularly Artificial Intelligence, instead of traditional methods. The overlap of Artificial Intelligence and Circular Economy can flourish many services in the environmental technology domain, in particular smart e-waste recycling, resulting in enabling circular smart cities.
OBJECTIVES: We analyse the growing need for automated e-waste recycling as an essential requirement to cope with the fast-growing e-waste stream and we shed the light on the impact of Artificial Intelligence in supporting the recycling process through smart classification of devices, where the smartphone is our case study.
METHODS: Our study applies transfer learning as a special technique of Artificial Intelligence by fine-tuning the output layers of AlexNet as a pre-trained model and perform the implementation on a small-size dataset that contains 12 classes from 6 smartphone brands.
RESULTS: We evaluate the performance of our model by tuning the learning rate, choosing the best optimizer, and augmenting the original dataset to avoid overfitting. We found that the optimizer of Stochastic Gradient Descent with Momentum and 3 $e^{-4}$ as a learning rate brings almost 98% model accuracy with generalization.
CONCLUSION: Our study supports automated e-waste recycling in decreasing the error-rate of e-waste sorting and investigates the advantages of applying transfer learning as the best scenario to overcome the rising challenges.

**Keywords:** Artificial Intelligence, Transfer Learning, Circular Economy, Automated E-Waste Recycling, Smart Cities.

## 1. Introduction

The enormous use of digitalization has a profound impact on every domain. Many concerns are raised with the growth of urbanization, like pollution, traffic congestion, rising welfare costs, and last but not least growing waste streams. The concept of Circular Economy (CE) was primarily aimed to enhance the recovery of end-of-life of products lifecycle by optimal recycling them, reusing them as raw materials, reducing the need to extract new resources, and closing the product loop. Smart Cities have been suggested as a solution to tackle the aforementioned problems, driven by digitalization, and to promote a sustainable environment through CE. In line with this, we discuss the role of digitalization, particularly Artificial Intelligence (AI), in the environmental technology domain and investigate how automated electrical and electronic waste (or the so-called e-waste) recycling can support shifting towards a sustainable environment, thus achieving CE goals. To achieve them efficiently, the classification of waste can maximize the performance of the whole process. Waste classification is a significant step for efficiently sorting and separating into different models and types. Therefore, the need for smart sorting is growing to support smart recycling.

The remainder of the paper is divided into the following sections: Section 1 introduces our research motivation, the importance of automated e-waste recycling driven by digitalization, and to achieve sustainable smart cities. Section

[*]Corresponding author. Email: nermeen.baker@hs-ruhrwest.de

2 reviews related work in automated waste classification. Section 3 discusses the background on Convolutional Neural Network (CNN) architecture. Section 4 presents our method. Section 5 is devoted to results and discussion and finally concludes our work.

## 1.1. Research motivation

In this article, we focus on adopting AI as a key technology for enabling automated e-waste recycling. One of the model services provided by the smart cities' concepts is digitalization, and due to the fact of the proliferation of digitalization, we considered e-waste as a model example. AI can be involved in many areas in the e-waste management system like collection, classification, sorting, etc. The motivation behind this study is gaining benefits of more proper e-waste recycling by automation, tackling the growing rate of e-waste in smart cities, and highlighting the advances of AI for this purpose. AI and CE can fuel the initiatives on smart cities to offer sustainable opportunities. As per [1], the biggest motivations for CE are technology development, new socio-economic opportunities, awareness, and change of consumer mentalities to adopt more sustainable products. On the contrary, some constraints face this movement, like poor legislation, lack of data collection standard process, and poor public participation. These obstacles inhibit smart cities from moving towards CE as well.

Our argumentation is to emphasize the importance of AI to encourage the automated recycling of e-waste in the smart cities' context by investigating the following aspects:

- The need to reduce human intervention by adopting automation and reducing the need for labor.
- Gain the benefits from applying AI techniques, particularly transfer learning, like using a small-size dataset rather than creating a big dataset, which is one of the painful tasks when designing a Neural Network (NN), decreasing the burden of long computational time, getting a high accuracy in sorting compared to the human-based process. Overall, our method supports reducing the error rate of e-waste sorting, and it is easier than building the NN from scratch.

To illustrate how we prove our investigation, we introduce the flow of the proposed method by insisting on the need for automated e-waste recycling and the impact of digitalization, transfer learning particularly on this process, to achieve circular smart cities as shown in Figure 1.

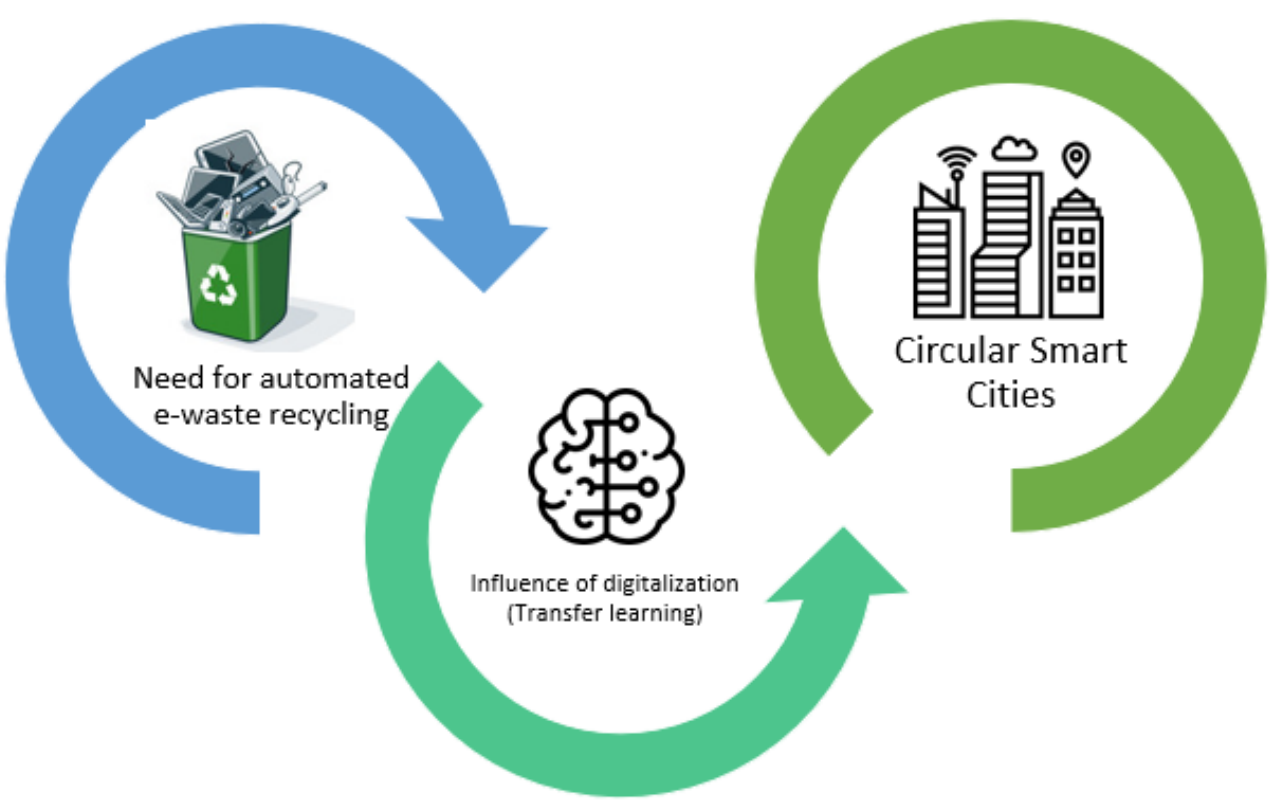

**Figure 1.** The flow of using transfer learning for automated e-waste recycling in smart cities

## 1.2. The need for automated e-waste recycling

In the take-make-dispose paradigm or the so-called linear economy, the generation of waste has dramatically increased in the last few decades. To overcome this problem, CE has been mainly proposed to confirm the social and environmental aspects of sustainability. Due to the great adoption of digitalization globally, e-waste is the fastest growing waste stream. This problem is one of the biggest challenges in reducing pollution, preserving valuable materials, and alleviating the toxicity and contamination of entering the eco-system. The Organization for Economic Cooperation and Development defined e-waste as "any appliance using electric power that is obsolete or has reached its end-of-life" [2].

The growing concern about informal e-waste recycling is alarming, especially on the improper way of processing in their final fate, like burning or melting them in acid baths and recovering only a few portions of valuable materials. General e-waste contains 60 different elements, like copper, aluminium, gold, platinum, and other metals represent 60% in e-waste, but 2.7% like cadmium, mercury, chromium are hazardous, and they will have poisonous and negative consequences on human health and the environment if they are not treated properly [3]. Moreover, when these recyclable elements are not recovered, new raw materials have to be extracted, and it will end in a lack of resources and higher energy consumption. As per [3], many Rare Earth Elements (REEs) are in e-waste, like 30% for silver in switches, 12% for gold in integrated circuits, 30% for copper in cables and 19% for cobalt in rechargeable batteries, and 79% for indium in Liquid-Crystal Display (LCDs), compared to mine global production. The previous figures aligned with growing sales of electronic devices and short lifespans, emphasize the initiatives to find a smart solution and adapting CE strategies like recycling e-waste.

Many directives tried to create legislations to process e-waste management; for example, the EU's Waste Electrical and Electronic Equipment (WEEE) stipulated the design of electronic devices should respect the eco-design, is easy to dismantle and recovered, and the producers are entitled to the take-back programs. Whereas in the US, the American e-waste recycling systems divided e-waste into ten categories based on the toxic substances, complexity, and profit from recycling [3].

The benefits of automated e-waste recycling, including a reduction in cost collection and complexity of e-waste processing, importing of raw materials, carbon dioxide (CO2) emission, and negative impacts on environment and labour health, besides boosting continuity of resources are the reasons behind choosing the need for automation of recycling to enable CE [4].

### 1.3. Influence of digitalization on automated e-waste recycling

Digitalization, including the Internet of Things (IoT), Big data, and AI, has a major influence on many sectors and can also be applied within the waste management system, resulting in improving the recycling process in two folds: for producers by enabling them to use recyclable materials and better purchasing and sorting decisions, as well as for recyclers for better waste sourcing options [5]. Consequently, these digital technologies act as catalysts to CE, but how can AI, in particular, mimic humans to create intelligent machines to solve such intricate problems?

AI plays a vital role in enhancing existing recycling infrastructure, including improving data collection and data mining processes to obtain a higher quality level than the typical analysing approaches besides automated sorting (where our study focus) leads to higher accuracy and better waste segregation quality. E-waste collection can also be improved by advances of AI, particularly in optimizing the e-waste collection routes to maximize the mass and the number of collected waste, besides navigation and tracking capabilities, especially the e-waste by storing, processing, analysing, and optimizing the necessary information, which ultimately will increase the whole waste management efficiency [6]. The next step is sorting e-waste, which is a prerequisite for high-rate recycling. The partnership between AI and robotics is gradually being adopted by many waste management applications, like analysing the streams of images and predicting the patterns to support the sorting process, and extending the lifespan of electronics through predictive maintenance [7].

### 1.4. Circular Smart Cities

We presented the term of circular smart cities in previous work [8], where the smart city paradigm could be linked to CE driven by the support of AI. The CE principles aligned with AI are expected to support the circular cities concept, and smart cities are consequently benefitting from implementing these developments [8].

Creating sustainable and eco-friendly cities with the help of digital technologies to provide smarter, liveable, and durable services is proposed under the label Smart City. The authors of [9], defined six characteristics of smart cities, which are: smart governance, smart mobility, smart living, smart people, smart economy, and smart environment. Our argumentation incorporates the last two components by increasing sustainable chances when adopting contemporary technologies.

On the other hand, the smart city concept may encounter some obstacles about the security and privacy issues, which drives much research to work on like, recognizing people' faces [10, 11] to access restricted areas [8–10], improving traffic flows by partly autonomous drones and vehicles [12] [13], traffic management and smart tracking, assistance systems [14, 15], predictive maintenance [16, 17], and last but not least, smart waste management [18].

The switch to digitalization can broadly improve the whole process of the waste management system by including digital identity tags for the waste container (IoT), digital order processing (Ecommerce), digital payments, digital communication with customers (chatbots), and storing and connecting with governmental databases (cloud services), which leads to better insights of waste patters (AI).

AI plays a fundamental role in supporting complicated services in many domains in the smart cities' context, due to the rising focus on digitalization-oriented technologies. The authors of [19] pointed out that new business models could be implemented by coherently applying AI as a successful potential for smart cities. This technology is essential due to the massive datasets gathered by sensors since this data needs to feed the decision support applications that leverage AI [20]. While the huge amount of data is created by different means in smart cities context, it needs to be turned into insights. AI is the key solution to play this role, in a various range of applications like healthcare, education, security, transportation, and the environment. AI has proved its ability to intelligently process large amounts of collected data created by sensors and produce significant information from it, based on recognizing patterns and features [21].

CNN is the basic building block of AI, and it has a special feature of self-programming with minimum human intervention, which gives AI to primarily act as a unique factor to enable circular smart cities. Further details about CNN architecture will be presented in section 3.

## 2. Related work

Automated recycling becomes an indispensable process due to the huge amount of the produced waste and its increasing detrimental effects on the environment and human health. It provides many advantages also to the economy.

One of the general classification models was introduced based on shapes, and dimensionality matching has been applied by [22]. It calculates the similarities and concurrencies of several shapes and uses the result to recognize the object.

Another classification method based on the reflectance properties of surfaces has been proposed by [23]. This method suggests an algorithm for estimation that learns correlations between surface reflectance and data enumerated from an observed image.

A study conducted by [24], who used the Bayesian framework with the help of the augmented Latent Dirichlet Allocation (aLDA) model to classify images based on their materialistic properties like glass, metal, fabric, etc. this classifier reaches 44.6% accuracy by processing the surface of each image.

Also, the waste domain classification using AI has been conducted in previous research. Using AI can be a very productive way to automate this process based on the collected images. A study by [25] used sensors in collaboration with machine learning algorithms to automatically sort waste based on textures and colours.

[26] used CNN to identify waste in images. The study presented a smartphone application to enable the user to report a pile of waste and identify its location, with an accuracy rate of 87%.

## 3. Background on CNN architecture

AI is a type of machine learning that allows the machine to develop using pattern recognition. There are three categories of learning: supervised, unsupervised, and reinforcement learning [27]. Our approach uses supervised learning to implement the classification using CNN, which is a NN that consists of layers as building blocks and uses the convolution operation in at least one of the layers, where the number of layers represents the depth of the network. If the number is large, the network will be called a deep neural network with many hidden layers. It can learn features directly from image data.

CNN is used to train a dataset divided into a training set, which is a set of images that corresponds to predefined classes (or labels), and a validation set to estimate the performance of the model. CNN uses the training set as labelled input-output pairs and performs training based on learning given examples, then it predicts the appropriate output for a given input. In order to obtain a good prediction performance, the training methods need to optimize the weights of each neuron-connection and to calculate the results to become as close as the expected class.

Most of CNN contains the convolution, pooling, fully connected, and softmax layers as building blocks, and they are defined as follows:

- CNN: The advantage of the convolutional layer, especially in classification, is that the kernel acts as a filter, sweeps the image in all directions, and catches the features by making the process shift-invariant. These filters are applied to each image to activate unique features (like edges, blobs, colours, brightness, etc.), then the output of each layer is used as an input to the consequent layer.
- Pooling layer: it is added usually between the consecutive convolutional layers. It simplifies the output by reducing the size of the matrices and gives a general look to the image, so the resulting matrix is smaller than the image matrix, but it contains the most prominent features. Usually, maximum or average functions are used for pooling in popular CNN architectures [28].
- Fully Connected (FC) layer: it is the last layer in CNN, which is used to flatten the 2D spatial features into a 1D vector and perform the learning.
- Softmax layer calculates the probability for each label in the dataset as an output of the model.

## 4. Method

To put the previous concepts, namely AI, CE, and circular smart cities together in practice, we used the following technical aspects.

Deep learning usually requires being trained on a huge amount of data on neural networks. To design a CNN from scratch, the network architecture should be well-designed, including the number of layers, the number and specification of filters, besides tuning the training parameters like learning rate, optimizers, and activation functions. Next, this network should be trained for a relatively long time on a huge dataset. A promising alternative to designing from scratch is using the transfer learning technique [29]. Transfer learning uses previously learned knowledge from a source task and transfers them to the target task. It is a special AI technique that helps a system adapt to new circumstances that allow processing data, extracting features, and making predictions. Pre-trained models are rich with feature representations because they were trained on a large number of images.

Our study uses transfer learning by freezing the transferred parameters from a pre-trained model, particularly AlexNet that was introduced by [30], which classifies one million high-resolution images (ImageNet) into one thousand labels. It is a deep convolutional neural network that consists of 650,000 neurons, 60 million parameters, and 630 million connections. Its architecture consists of eight layers, including five convolutional layers, and three fully connected layers, besides three pooling operations, as shown in Figure 3 for the original model. In standard AlexNet architecture, the first two convolutional layers are followed by an overlapping max-pooling layer, the other three convolutional layers are connected directly, and the final convolutional layer is followed by a max-pooling layer. AlexNet has been used extensively in the research due to its simple, and not-so-deep architecture. One of the main characteristics of AlexNet is using Rectified Linear Unit (ReLU) activation function that leads to faster training than other activation functions like sigmoid or tanh. It is an effective activation function that maps the negative values with zeros and maintains positive values. Another advantage of using AlexNet that it has a dropout layer. CNN has a huge number of parameters that can cause overfitting, which can be prevented by regulating the network to memorize them too much. Practically, it can be implemented by randomly stopping the neuron's contribution in forward or backward propagation, leading to dropping the units with their connection during training [31].

For implementation, we created the dataset manually from the Internet. It contains 650 RGB images. The computing environment was Matlab R2020b with a deep learning toolbox was used for implementation installed on a laptop used Windows 10 (64 bits) equipped with i5 processor, and requires Matlab parallel computing toolbox with CUDA (which is a parallel computing platform and programming model developed by NVIDIA for general computing on a Graphics Processing Unit (GPU)) of ASUS Nvidia GeForce RTX 2070S 8GB for the acceleration of training process.

The dataset consists of 12 smartphone models, a relatively small dataset, from 6 brands, namely Acer, HTC, Huawei, Apple, LG, and Samsung. Since most of the frontside of smartphones look similar recently, we collected images that focus on the backside where unique features like the logo, camera lenses are distinguished. The dataset is split into 80% for the training set and 20% for the validation set. Figure 2 shows an example of a subset of the dataset.

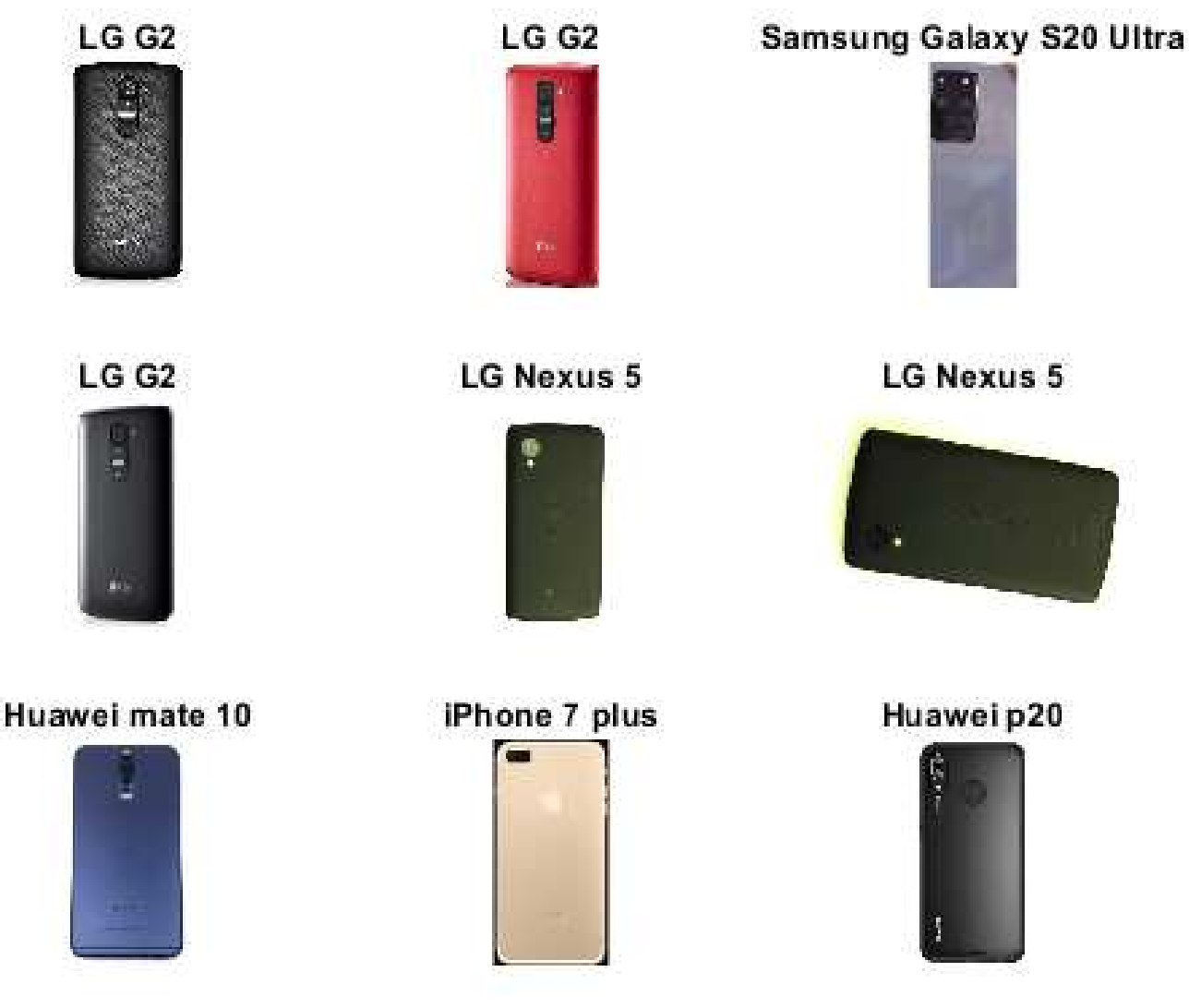

**Figure 2.** Example of a subset of the dataset

We started by loading AlexNet, replacing the last three layers to classify 12 labels, then training the network on our smartphone dataset, finally assessing the network on the validation set, and checking the performance.

Regarding the training options, we set the mini-batch size to 64, which represents the number of the subsets of the training set that are processed on GPU simultaneously. After the whole batch is sent to the network and the error of the batch is propagated backward into the weights, every weight in the network is being updated. Higher values of batch-size lead to better convergence and higher accuracy. However, it is limited to the available memory of the GPU [25]

Full pass of training process over the entire training set uses mini-batches called one epoch. To control the early stop, we set the max epochs as 30, and the training set was shuffled before each epoch.

In the beginning, the network was initialized with frozen pre-trained weights for all layers except the last three layers, as described in previous work [8]. The learnable weights of AlexNet are frozen in the fully connected layer. To perform the fine-tuning, replacing this layer with a new fully connected layer has an output value equal to the number of classes in the new task.

The details of the proposed implementation are presented in Figure 3.

## 5. Results and discussion

After setting the layers configuration and the training options, the model is ready for prediction. Evaluating the performance of the network is a challenging task and depends on the computational complexity. We performed three different experiments to choose the best parameters for our model with a baseline of data augmentation, optimizers, and learning rate.

### 5.1. Baseline: data augmentation

Data augmentation is creating alternative copies of the original dataset by adding more images effortlessly, and it is mainly used to alleviate small-size datasets and overfitting problems.

A significant factor that should be evaluated is a generalization. If there is a large distance between the training and validation accuracy, practically happens when the model is very complex for the available amount of the training set, and the model is not able to generalize, or the so-called overfitting.

The following operations were applied to perform data augmentation, random X reflection, random Y reflection, random X translation, random Y translation, random X scale, random Y scale, random X shear, random Y shear, where the scale range is [0.9 1.1] and the translation and shear range is [-50 50] pixel. An example of our dataset augmentation can be seen in figure 4. As a result, each image is multiplied by 9 to get 5850 images in the dataset, including the original unchanged set.

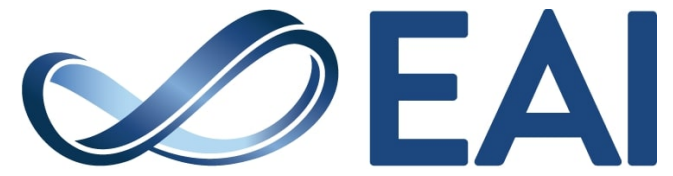

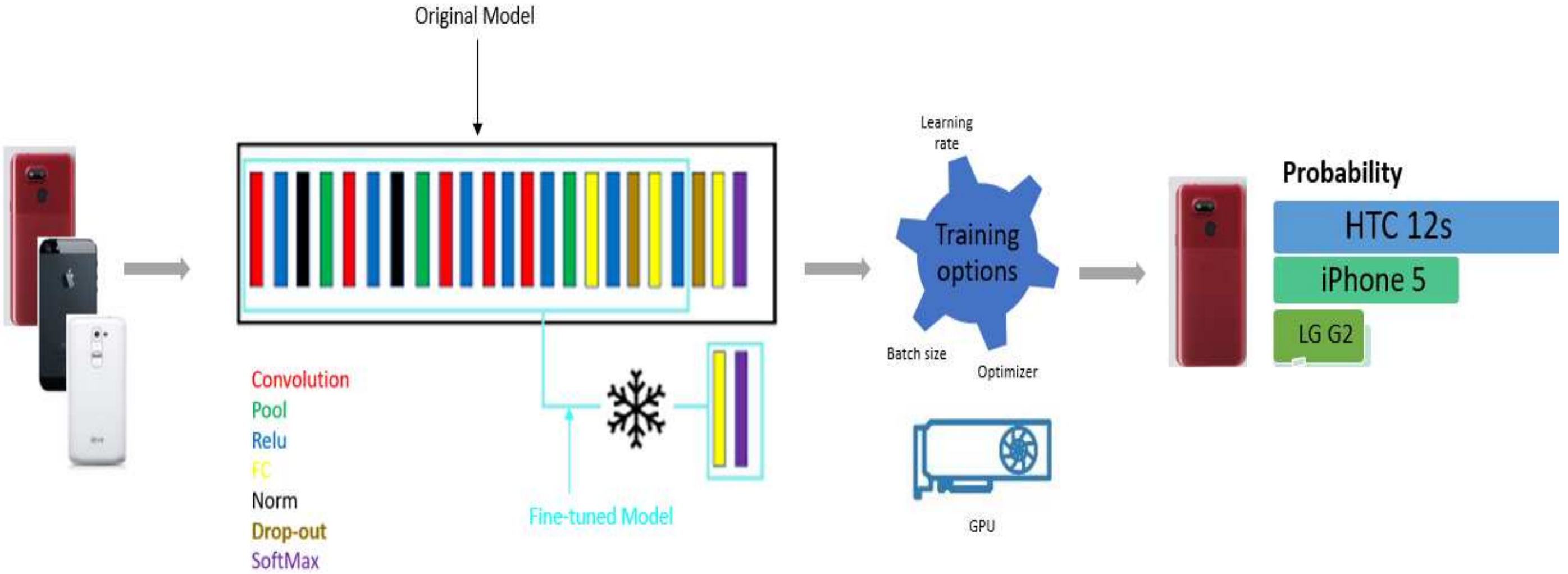

**Figure 3.** The implementation diagram of the original AlexNet architecture and the fine-tuned model

## 5.2. Baseline: Optimizers

Choosing the right optimizer can help to reach the global minima, reduce the loss function, and set up the correct parameters. The loss function is one of the most important metrics for testing the network performance, which represents the difference between the predicted output and the target class. To minimize the loss function, the gradient descent updates the weights and biases of the network by taking small steps at each iteration of the opposite direction of the gradient. To set the training options, we tested three popular optimizers.

Gradient descent considers the whole data at one time that leads to redundant and inefficient computation, but Stochastic Gradient Descent (SGD) computes random selection or small subset instead. However, SGD may oscillate along the path of steepest descent towards the optimum, where the surface curves have more steeply on the dimension. The momentum alternatively helps to accelerate SGD towards the local minima and reduce oscillations. SGD with momentum, or SGDM, uses a single learning rate for all parameters, whereas RMSProp (which is a gradient-based optimization technique used in training neural networks) tries to improve the network performance by adapting learning rates by parameter to optimize the loss function [32]. In comparison, the adaptive learning rate optimization algorithm (Adam) computes individual adaptive learning rates and momentum to converge faster. It uses an estimation of the first and second moments of the gradient to adapt the learning rate for each weight [33]. We performed seven learning trials and Table 1 shows the accuracy range of each method.

Table 1. The accuracy mean and standard deviation of each method.

| Implementations 3e-4 | Accuracy |
|---|---|
| **SGDM** | 98.3505 ± 2.0155 |
| ADAM | 95.2577 ± 3.0427 |
| RMSProp | 93.1959 ± 3.8635 |

From the table, it is clearly noted that SGDM provides better performance compared to ADAM and RMSProp. This result also supports the study presented in [34], who conducted empirical research and stated that, although Adam proves that it converges faster than other optimizers, it does not converge to the optimum solution and generalize well in classification as SGDM does.

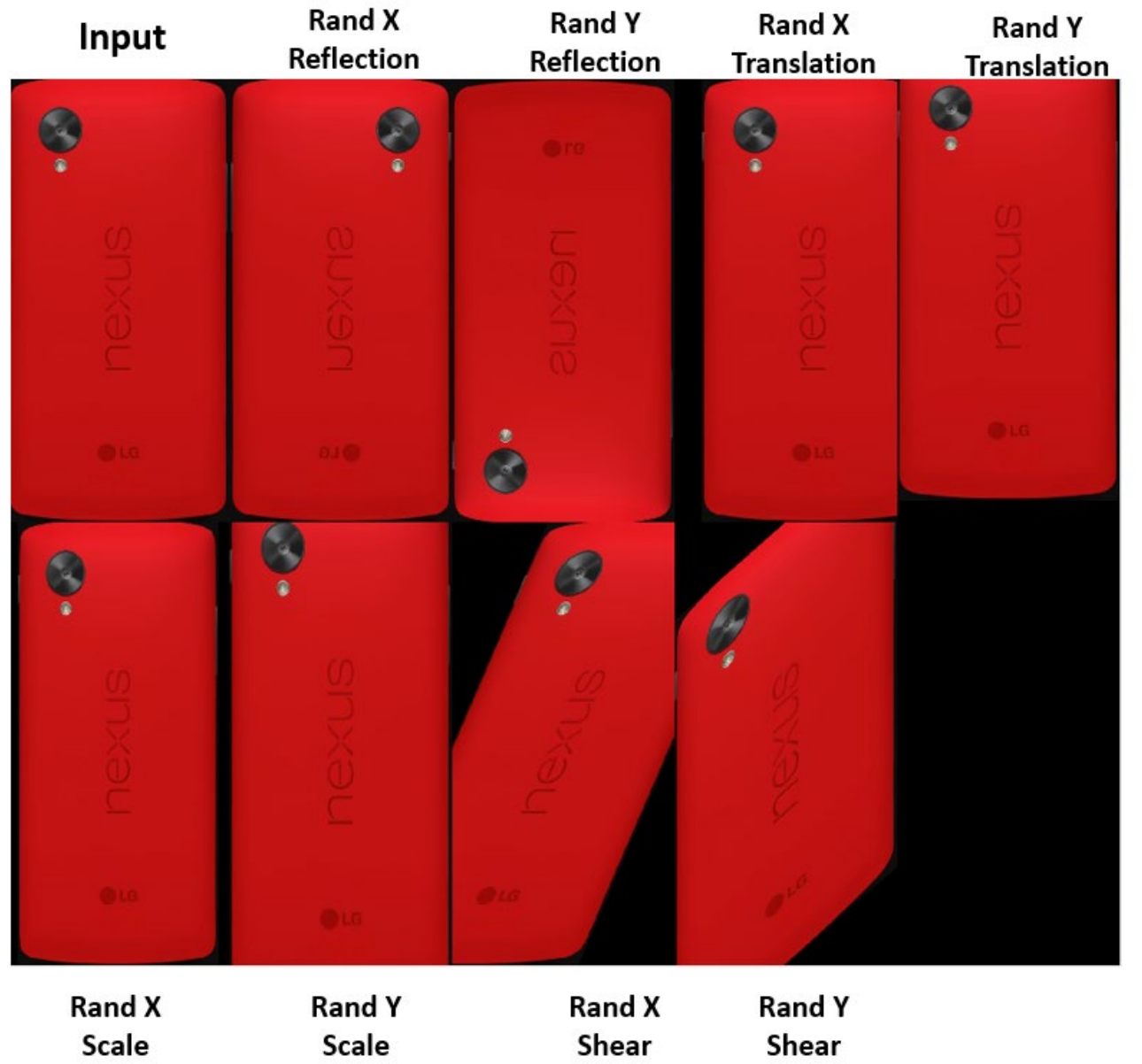

**Figure 4.** Example of used data augmentation

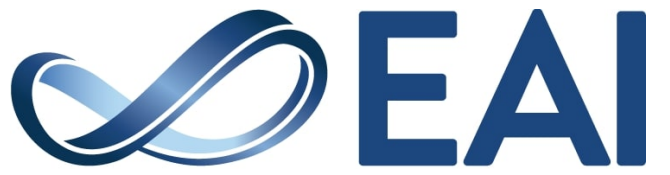

## 5.3. Baseline: learning rate

Learning rate controls the speed of training, for a smaller learning rate the model could have higher accuracy, but it takes a longer time to train. One of the biggest challenges of gradient descent is choosing a proper learning rate. Too small learning rates may dwindle around the minimum and get slow convergence and too large learning may cause unstable training process. Transfer learning mostly uses a smaller learning rate since the learned weights have already significant optimization.

Performance comparison based on a variety of learning rates has also been conducted to choose the best option for the proposed model. Figure 5 visualizes a box chart that represents the distribution of accuracies. The median accuracies per each box are drawn in the middle of the box, and the upper and lower quartiles are shown at the top and the bottom edges of the box, respectively. The whisker endpoints illustrate the lowest and highest accuracy.

We tried to guess the learning rate by reducing the learning rate when the loss oscillates widely and keeps getting worse, whereas when the loss is slowly and consistently falling we increased the learning rate.

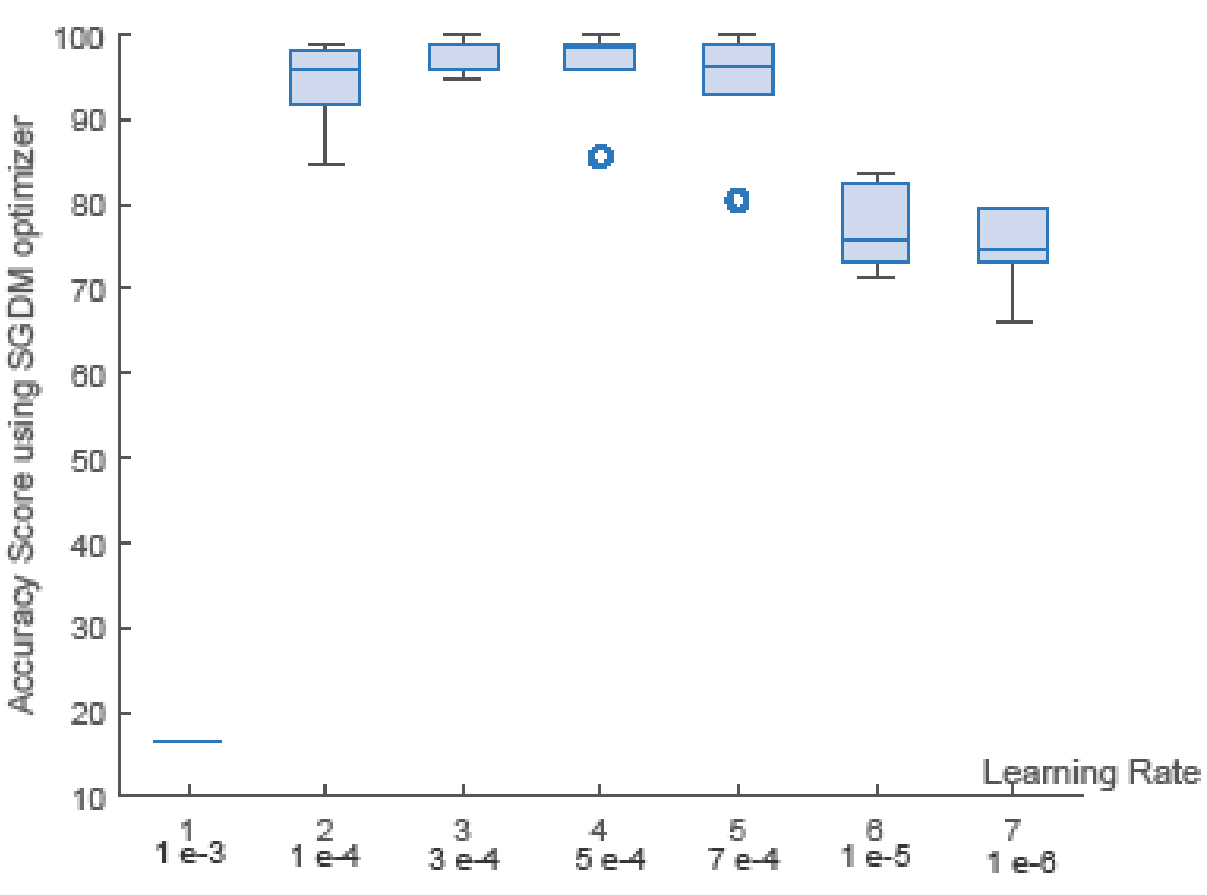

**Figure 5.** Box chart of Accuracy Scores for different learning rates using SGDM optimizer

From the box chart, it is clearly illustrated that the learning rate that corresponds to $3\,e^{-4}$ provides better performance compared to other learning rates.

As shown in Figure 6, a combination of setting the proposed data augmentation and SGDM optimizer with $3\,e^{-4}$ learning rate has the best model generalization performance and reaches almost 98% accuracy.

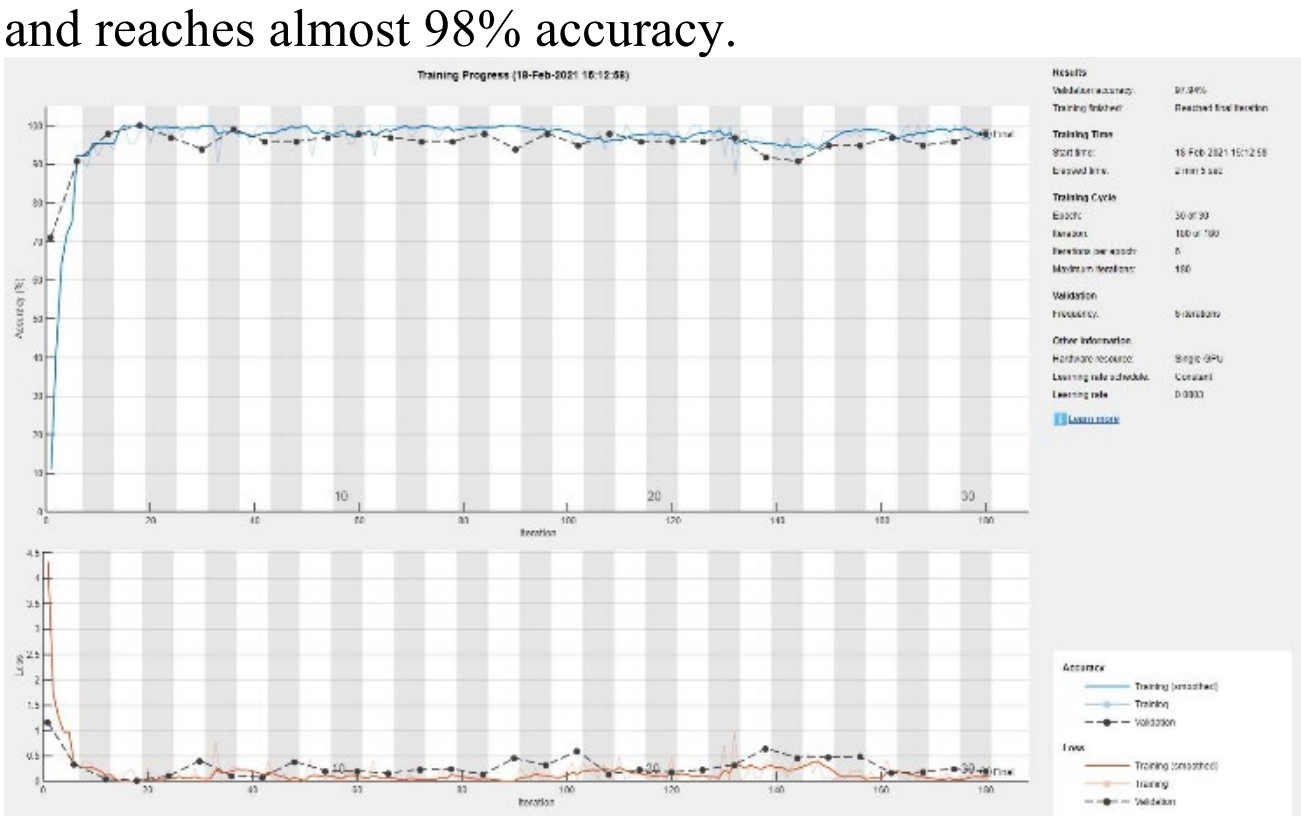

**Figure 6.** Accuracy and loss function performance of our model

## 6. Conclusion

In a nutshell, our method underlines the important role of AI in shifting towards automated e-waste classification, hence supporting circular smart cities. By the example of AI-enhanced automatic smartphone classification, we showed that e-waste management could be significantly enhanced by using digital technologies that speed up the process. The suggested method supports two important decision factors in implementation, by reducing the error rate of e-waste sorting and it is easier to use transfer learning than building the NN from scratch.

We tested the performance by the tuning learning rate and optimizer, besides performing data augmentation to avoid the overfitting and small-size dataset problems.

However, our approach should not only be used as an end-of-pipe technology, which may result, for example, in so-called rebound-effects. Hence, it cannot replace the transition to a more sustainable economy and society, which requires dedicated efforts in the respective fields of action, such as new business models or user preferences. Nevertheless, our approach and automated e-waste management, in general, could alleviate contemporary e-waste problems quickly and diminish cost effectively, which builds a basis for a more sustainable world in the future. Automated e-waste classification is not the end of the story. Therefore, we try to develop a (semi) automated e-waste management system in our Circular Digital Economy Lab (CDEL) in order to get the full benefits from CE and digitalization, where we integrate our model with other systems such as robotics, IoT, and data mining.

**Acknowledgements**

This work has been funded by the Ministry of Economy, Innovation, Digitization, and Energy of the State of North Rhine-Westphalia within the project Prosperkolleg.

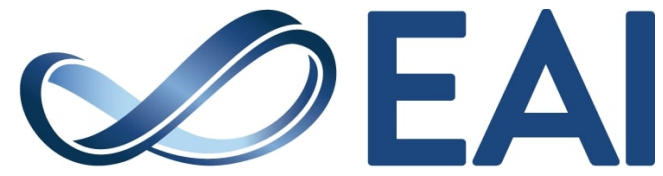

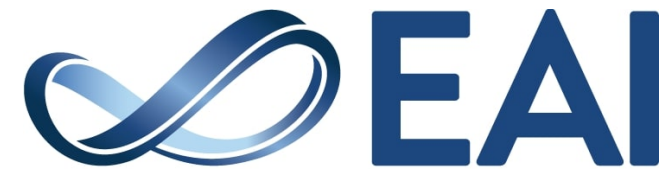